\documentclass[conference,compsoc]{IEEEtran}

\usepackage{graphicx}
\usepackage{float}
\usepackage{amsmath} 
\usepackage{amssymb}
\usepackage{mathtools}
\usepackage{amsfonts}
\usepackage{algorithmic}
\usepackage{stfloats}
\usepackage{url}

\usepackage[backend=biber, style=numeric, sorting=none]{biblatex}
\begin{document}

\title{Concept: Dynamic Risk Assessment for AI-Controlled Robotic Systems}

\author{\IEEEauthorblockN{Philipp Grimmeisen, Friedrich Sautter, and Andrey Morozov}
\IEEEauthorblockA{Institute of Industrial Automation and \\Software Engineering\\
University of Stuttgart\\
\{first\}.\{last\}@ias.uni-stuttgart.de}}

\maketitle

\begin{abstract}


AI-controlled robotic systems pose a risk to human workers and the environment. Classical risk assessment methods cannot adequately describe such black box systems. Therefore, new methods for a dynamic risk assessment of such AI-controlled systems are required. 
In this paper, we introduce the concept of a new dynamic risk assessment approach for AI-controlled robotic systems. The approach pipelines five blocks: (i) a Data Logging that logs the data of the given simulation, (ii) a Skill Detection that automatically detects the executed skills with a deep learning technique, (iii) a Behavioral Analysis that creates the behavioral profile of the robotic systems, (iv) a Risk Model Generation that automatically transforms the behavioral profile and risk data containing the failure probabilities of robotic hardware components into advanced hybrid risk models, and (v) Risk Model Solvers for the numerical evaluation of the generated hybrid risk models.


\end{abstract}

\IEEEpeerreviewmaketitle

\section{Introduction}
\begin{figure}[h]
    \centering
    \includegraphics[width=0.4\textwidth]{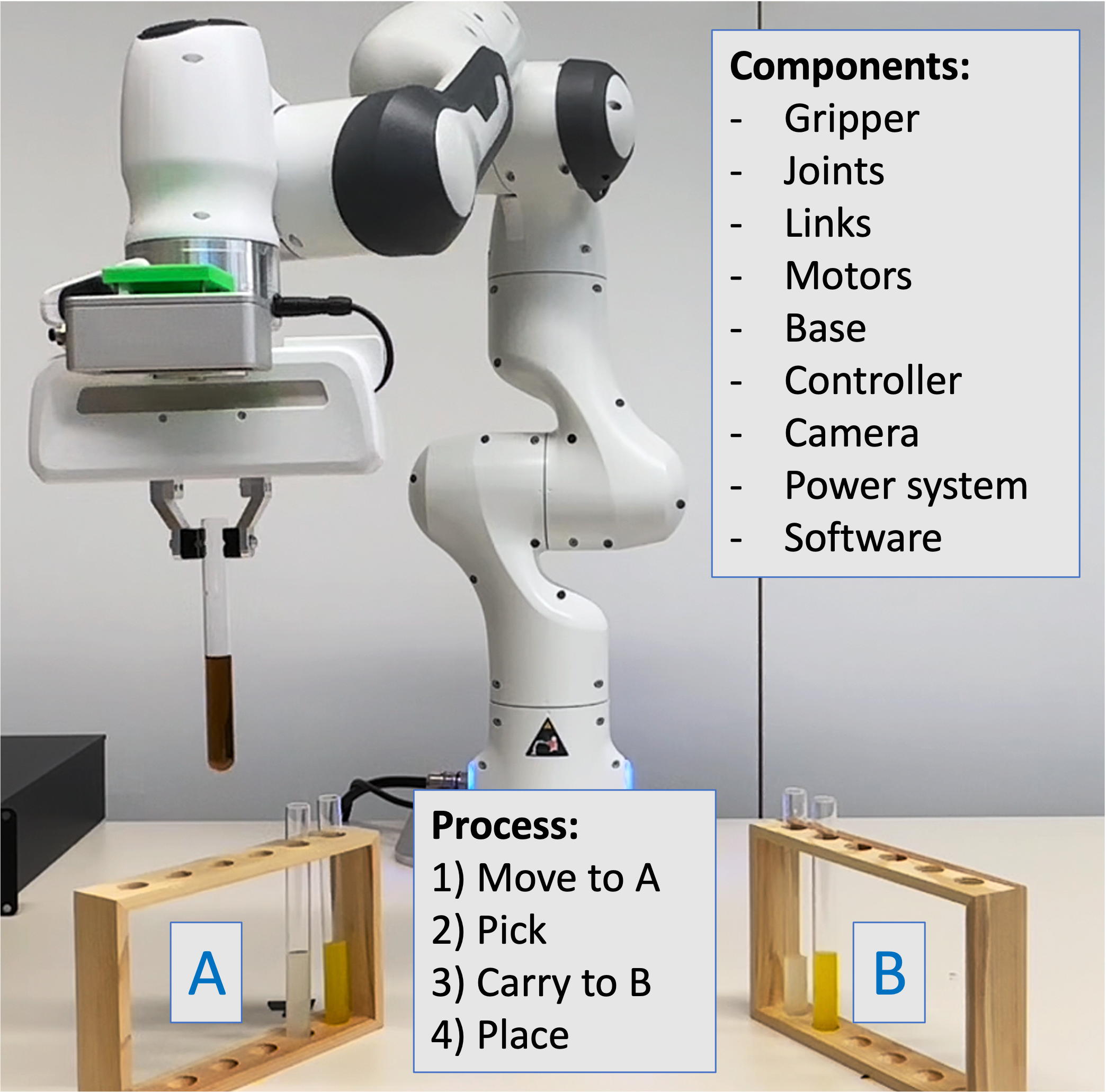}
    \caption{Robotic manipulator which has been trained to grasp a test tube at position A and place it to position B.}
    \label{fig:Franka}
\end{figure}

Robotic systems can be safety-critical because they are often deployed in close proximity to human workers. Moreover, they collaborate with human workers. They may contain moving parts that can pose risks to human workers or carry substances dangerous to humans, such as chemical mixtures, as shown in Figure \ref{fig:Franka}.


Robotic systems are increasingly being controlled by AI algorithms such as Reinforcement Learning \cite{ibarz2021train}. This requires a shift of the paradigm of risk assessment towards a dynamic risk assessment instead of the classical approach when the risk assessment of the system is done only once before the operation. The research question to be answered is: How to automatically and dynamically estimate the potential risk of robotic systems controlled by black-box policies? 



\begin{figure*}[thbp]
    \centering
    \includegraphics[width=0.95\textwidth]{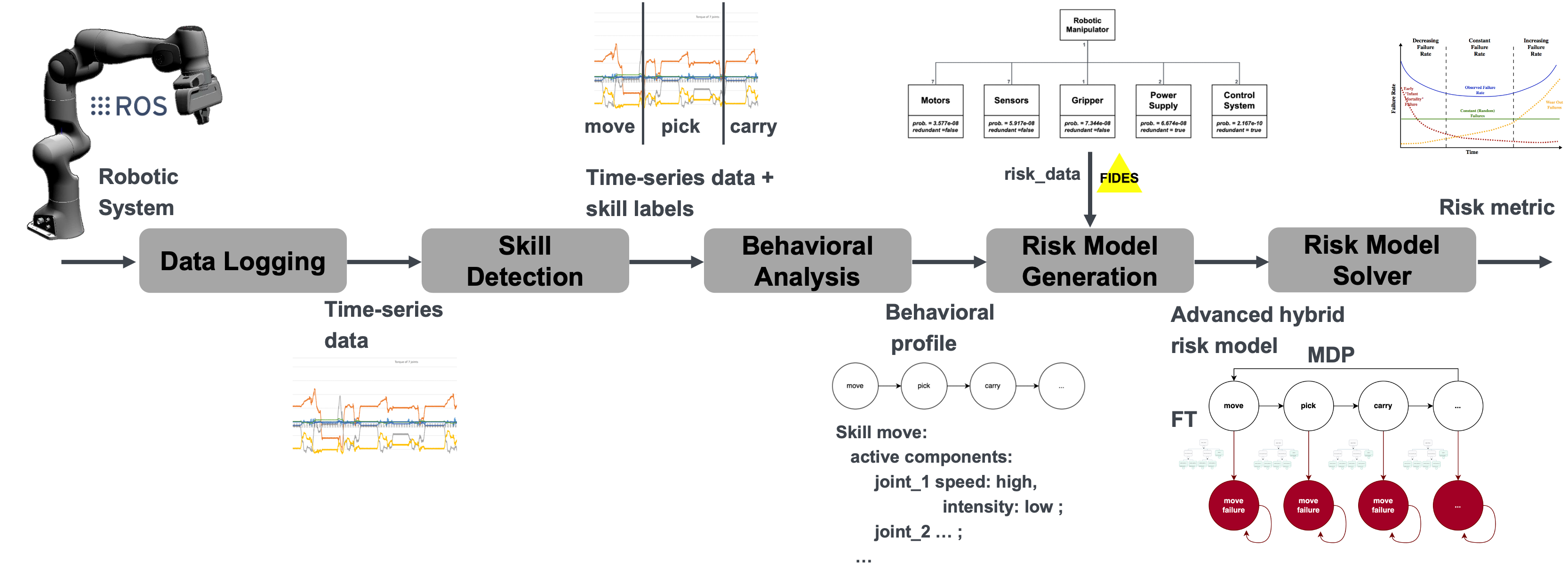}
    \caption{The proposed approach to the dynamic risk assessment of AI-controlled robotic systems.}
    \label{fig:overview}
\end{figure*}

\textbf{Contribution:} This paper presents a concept of a new approach that enables the dynamic risk assessment of unknown or dynamic control policies including AI-controlled robotic systems. The proposed pipeline, shown in Figure \ref{fig:overview}, consists of five blocks:
\begin{itemize}
    \item Data Logging consists of a hardware-interface integrated ROS logger. This block analyzes the in-simulation executed ROS code. It logs time-series data such as the positions and velocities of each joint.
    \item Skill Detection block is a deep learning-based classifier. This classifier detects the executed skills and labels the time-series data with these skills e.g. move, pick, carry, etc. 
    \item Behavioral Analysis block generates the behavioral profile of the given robotic systems. The behavioral profile contains information about the sequence of executed skills, active components for each skill, and properties of components such as velocity, execution time, intensity, etc. that influence reliability characteristics of the components.
    \item Risk Model Generation block transforms the behavioral profile and risk data, e.g. failure probabilities of robotic hardware components, into advanced hybrid risk models.
    \item Finally, the Risk Model Solver quantifies the generated risk models.
\end{itemize}

\section{Concept}

The robotic manipulator as shown in Figure \ref{fig:Franka} has been trained to grasp a test tube at position A and place it to position B. The robotic manipulator uses the skills Move to A, Pick, Carry to B, and Place. The real robotic manipulator consists of several components such as Gripper, Joints, Links, Motors, Base, Controller, Camera, Power system, and Software.

\subsection{Data Logging}
 The input of the Data Logging is a Gazebo simulation that executes a control policy, in the form of any ROS code that utilizes the hardware abstractions. The control policy could be the policy that controls the robotic manipulator to grasp a test tube at position A and place it to position B, as shown in Figure \ref{fig:Franka}.
 
 The hardware-interface integrated ROS logger publishes messages about active robotic components. These ROS topics in Gazebo, include among others, information about the position, velocity, effort, and execution time of robotic components. More specifically, we log the positions and speeds of each joint. The log contains all this information as time-series data. The logger can log data from any ROS code independent of the programming language (Python or C++) and ROS version (ROS1 or ROS2). The logger is not hardware-specific and can log the data of any ROS-controlled robotic policy. To map the hardware components directly to risk data, the logger demands hardware descriptions to be published inside the existing hardware abstractions.




\subsection{Skill Detection}
The deep learning-based skill detector developed for robotic manipulators is one of the key elements of this approach. It labels the logged time-series data with executed skills. Executed skills could be move, pick, carry, etc. The model shall be trained on meticulously self-generated data sets, including training and test data. Therefore, the skill detection can reliably detect and classify robot skills. We generate the data set from various robotic control algorithms. We arbitrarily choose the parameters, such as speed, execution times, path planning, etc. That gives us a particularly comprehensive data set. Preliminary the detector should combine Long Short-Term Memory (LSTM), Multilayer Perceptron (MLP), and possibly attention based architectures.

\subsection{Behavioral Analysis}
The Behavioral Analysis block interprets the time-series data with skill labels. To interpret the time-series data, we will use a rule-based approach. 
The Behavioral Analysis block creates the sequence of executed skills with exact time periods. In addition, it maps active components to the corresponding skills and adds available properties to the component. The Behavioral Analysis creates and outputs the behavioral profile of the system. It contains a list of skills (e.g. move, pick, carry, place, etc.) with exact execution time, a list of active components (joints, torque sensors, camera, gripper, etc.) during each skill, and the corresponding properties e.g., velocity, active time, effort, etc. for each component. The properties of each component can be different for each skill.

\subsection{Risk Model Generation}
The Risk Model Generation block parses the behavioral profile and risk data. The risk data contains failure probabilities, redundancy information, and other fault tolerance and resilience mechanisms of robotic hardware components. This information has to be added by experts. 
The transformation algorithm creates a fault tree \cite{ruijters2015fault} for each detected skill. Basic events of the fault tree are the failures of system components that are active during the execution of the skill. 
The logic of the fault tree represents potential fault tolerance mechanisms such as redundancies, recoveries, spares, etc. In case of no fault tolerance the top event is the failure of the skill modeled as an OR-gate. If the algorithm detects a redundant component, it creates an AND-gate and adds the corresponding quantity of the components as basic events. The failure probabilities of the basic events can vary depending on the logged execution time, velocity, intensity, etc. Reliability block diagrams \cite{kim2011reliability} could be used as an alternative to fault trees. In case of dependent failures, fault tree models can be combined with Bayesian networks \cite{bobbio2001improving}. 

According to the sequence of executed skills, the transformation algorithm creates a dynamic high level risk model based on a Markov chain \cite{fuqua2003applicability}, stochastic Petri net \cite{bause2002stochastic}, dual graph error propagation model \cite{morozov2011dual}, or PRISM language \cite{kwiatkowska2002prism}. In case of a simple discrete-time Markov chain model, it creates a transient state for each skill and a corresponding absorbing failure state. An absorbing "done" state is created, which serves as a state of success. The transition probability from a state to its corresponding failure state is defined by interconnected fault trees. The output of the transformation algorithm is a hybrid risk model that contains a Markov chain with interconnected fault trees in the OpenPRA model exchange format.


\subsection{Risk Model Solver}
We use the OpenPRA framework \cite{grimmeisen} for numerical risk assessment. OpenPRA is an open-source framework, which integrates multiple risk methods into an easy-to-use environment. The OpenPRA integrated analysis module can analyze hybrid risk models, such as Markov chains with interconnected fault trees. It first solves the fault trees by calling the fault tree analysis solver. The computed results are added to the transitions of the Markov chain. The Markov chain solver solves afterward the final Markov chain.

\section{Conclusion}
In this paper, we introduce a concept of a new approach, that enables the dynamic risk assessment of dynamic or unknown control policies including AI-controlled robotic systems. The novel approach includes five key methods: Data Logging, Skill Detection, Behavioral Analysis, Risk Model Generation, and Risk Model Solver. 

The approach primarily addresses classical hardware failure. However, we are currently working on an approach that uses large language models to automatically find failure modes and analyze these failure modes with reinforcement learning. The reinforcement learning-based analysis will allow us to understand which events or sequences of events lead to certain failure modes. Events include faults (e.g. bit-flip or noise) that will be injected into the system with fault injection methods. The approach will help us to assess the risk of the software parts of the system more precisely. We are working on the realization of this approach in parallel. 
\printbibliography

\end{document}